\documentclass[conference]{IEEEtran}
\IEEEoverridecommandlockouts
\usepackage{cite}
\usepackage{amsmath,amssymb,amsfonts}
\usepackage{algorithmic}
\usepackage{graphicx}
\usepackage{textcomp}
\usepackage{xcolor}

\graphicspath{ {./figures/} }
\usepackage{hyperref}
\usepackage{float}
\usepackage{verbatim} %comments
\usepackage{soul}
\usepackage{multirow}
\usepackage{subcaption}
\sethlcolor{yellow}
\restylefloat{figure}
\restylefloat{table}

\def\BibTeX{{\rm B\kern-.05em{\sc i\kern-.025em b}\kern-.08em
    T\kern-.1667em\lower.7ex\hbox{E}\kern-.125emX}}
\begin{document}

\title{Using machine learning for fault detection in lighthouse light sensors
}

% \author{\IEEEauthorblockN{Anonymous Authors}}

\author{\IEEEauthorblockN{Michael Kampouridis}
\IEEEauthorblockA{\textit{School of Computer Science and Electronic Engineering} \\
\textit{University of Essex}\\
Wivenhoe Park, UK\\
mkampo@essex.ac.uk}
\and
\IEEEauthorblockN{Nikolaos Vastardis}
\IEEEauthorblockA{\textit{General Lighthouse Authorities of the UK and Ireland}\\
\textit{Research and Development Directorate}\\
Harwich, UK\\
Nikolaos.Vastardis@gla-rad.org
}
\and
\IEEEauthorblockN{George Rayment}
\IEEEauthorblockA{\textit{School of Computer Science and Electronic Engineering} \\
\textit{University of Essex}\\
Wivenhoe Park, UK\\
gr17754@essex.ac.uk
}
}

\maketitle

\begin{abstract}
Lighthouses play a crucial role in ensuring maritime safety by signaling hazardous areas such as dangerous coastlines, shoals, reefs, and rocks, along with aiding harbor entries and aerial navigation. This is achieved through the use of photoresistor sensors that activate or deactivate based on the time of day. However, a significant issue is the potential malfunction of these sensors, leading to the gradual misalignment of the light's operational timing. This paper introduces an innovative machine learning-based approach for automatically detecting such malfunctions. We evaluate four distinct algorithms: decision trees, random forest, extreme gradient boosting, and multi-layer perceptron. Our findings indicate that the multi-layer perceptron is the most effective, capable of detecting timing discrepancies as small as 10-15 minutes. This accuracy makes it a highly efficient tool for automating the detection of faults in lighthouse light sensors.
\end{abstract}


\section{Introduction}
\label{introduction}

The General Lighthouse Authority (GLA) of the UK and Ireland is dedicated to providing a reliable, efficient, and cost-effective navigation aid service for the maritime community's safety and benefit. The GLA Research and Development (GRAD) division, serving all three General Lighthouse Authorities in the UK and Ireland, is at the forefront of this mission. GRAD is responsible for researching and developing both physical and radio marine aids to navigation (AtoNs), as well as supporting systems and their integration, to uphold the GLA's commitment to delivering top-notch AtoNs for the safety and benefit of mariners.

GRAD manages a variety of AtoNs, including lighthouses, buoys, light-vessels, beacons, and electronic navigation systems. These aids are crucial for safely guiding mariners through some of the UK's most trafficked waters, like the Dover Strait, the world's busiest shipping lane.

Among these aids, lighthouses play a vital role in marking perilous coastlines, shoals, reefs, rocks, and assisting in both sea and aerial navigation. A critical component of a lighthouse is its photoresistor sensor, which automates the light's operation. However, sensor malfunctions pose significant risks, such as delayed activation of lights, endangering ships in the vicinity by not alerting them to nearby hazards.

The challenge lies in monitoring and addressing sensor malfunctions in lighthouses, often situated in remote and hard-to-access locations. Reaching these sites, sometimes requiring costly helicopter transport, drives up maintenance expenses significantly. As a result, it's more feasible to replace unreliable components during regular maintenance visits rather than immediately upon detecting a fault.

To proactively detect potential sensor faults and efficiently plan maintenance, we propose utilizing machine learning (ML) to identify early signs of malfunction in lighthouse light sensors. A major hurdle is the absence of historical data on verified sensor failures, largely due to preemptive replacements during scheduled maintenance. Sensor failures can be gradual, showing increasing delays in light activation, or abrupt, such as a total breakdown. Our focus is on the former, the more complex scenario, as abrupt failures are straightforward and don't require ML intervention.

To overcome the data limitation, we simulate gradual photoresistor sensor faults and apply pre-trained ML models to this data. This approach helps us evaluate how effectively these models can detect declines in performance metrics like accuracy and F1 score. We also explore the models' capability to swiftly pinpoint potential faults. Our analysis includes four prominent algorithms: decision trees, random forest, extreme gradient boosting, and multi-layer perceptron. The goal is to develop a system that enables early detection of light sensor faults, aiding GRAD in prioritizing lighthouse maintenance.

The structure of this paper is as follows: Section \ref{sec:review} reviews existing literature and sets the context for AtoNs, highlighting the scarcity of ML applications in this domain. Section \ref{sec:methodology} details our methodology, while Section \ref{sec:setup} describes the experimental setup. Section \ref{sec:results} discusses our findings, and finally, Section \ref{sec:conclusion} concludes the paper and suggests directions for future research.

\section{Background information and literature review}\label{sec:review}

Machine learning has garnered considerable attention across various sectors, including economics \cite{kampouridis2012market}, finance \cite{adegboye2021machine}, autonomous vehicles \cite{lee2019maturity}, smart homes \cite{vastardis2016user}, facial and image recognition \cite{zhao2003face,litjens2017survey}, telecommunications \cite{shaghaghi2013guided}, weather \cite{cramer2017extensive}, and object detection \cite{vahab2019applications}.

Despite its widespread application, the realm of Aids to Navigation (AtoN), such as lighthouses, remains largely untouched by automation, particularly in maintenance practices. Current literature does not encompass the use of automated methods like machine learning for detecting AtoN failures. However, there is a substantial body of work on fault detection in various maritime industry aspects, including emissions monitoring \cite{CAPEZZA2019375}, risk assessment in gas turbine systems on specialized tankers \cite{AHN2017226}, analysis of risks in ship mooring operations \cite{CEMKUZU2019128}, maintenance prioritization in ship systems \cite{dikis2014probabilistic}, predictive maintenance of marine main engines \cite{dikis2019dynamic}, exhaust gas valve monitoring \cite{Fog1999554}, condition monitoring in marine engines \cite{BASURKO2015404}, diagnosis of engine cylinder covers \cite{4459891}, naval vessel system decay studies \cite{CIPOLLINI201812}, and damaged mooring equipment detection \cite{BEGG2018577}. Beyond the maritime sector, machine learning has been applied in detecting failures in agricultural machinery \cite{rajakumar2021health}, wind turbines \cite{kusiak2011data}, aircraft components \cite{savitha2020online}, and in production plants \cite{kolokas2020generic}. Failure identification encompasses various aspects, including detection, diagnosis, condition monitoring, prediction, and forecasting of variables like remaining useful life or degradation \cite{leukel2021adoption}.

A notable challenge in failure detection is the scarcity of labeled data indicating failures. While there's ample data on functional vessel components, data showcasing failures are rare. To overcome this, researchers often resort to creating \textit{simulated data} for various fault scenarios \cite{cheliotis2020machine,cipollini2018condition,coraddu2016machine}.

In our study, we encounter a similar challenge due to the lack of labeled features signifying failures. Although there's an abundance of historical data on photoresistor sensor activity (e.g., times when a light sensor was activated or deactivated), instances of sensor failures are minimal. This scarcity is attributed to the GLA's preventive maintenance approach during scheduled engineer visits. Due to the insufficiency of historical data on sensor failures, training a machine learning model on existing data to predict future failures is not feasible. To address this, we simulate various sensor failures and examine the response of pre-trained machine learning models to these scenarios, aiming to identify potential faults. The following section details this process and outlines the remainder of our methodology.

\section{Methodology}\label{sec:methodology}

Our research aims to enable the early detection of faulty sensors, a crucial step in ensuring timely replacement by the GLA maintenance team. Timely identification of such faults is not only critical for navigational safety, but also aids in efficient planning. Given that lighthouses are distributed across the UK, often in remote and hard-to-access locations, pinpointing faulty sensors can significantly enhance the logistics of maintenance trips, thereby reducing associated costs.

We tackle this challenge by framing it as a classification problem\footnote{Initial explorations included the use of clustering algorithms to segregate sensor data into normal and abnormal behavior categories. However, classification approaches yielded more promising results, leading us to concentrate our experiments in this area rather than clustering.}. In this setup, we aim to predict the operational status of lighthouse lights (on or off) by analyzing a set of climate-related features. This predictive approach not only assists in the immediate identification of malfunctioning sensors but also contributes to more strategic, cost-effective maintenance planning across the network of lighthouses.

Confronted with the absence of labeled fault instances in our sensor data logs, we have adopted an alternative approach. Our first step involves training various machine learning (ML) algorithms on historical data. The objectives are twofold: (i) to determine the most effective algorithm for our needs, and (ii) to establish a baseline of what constitutes `normal' behavior for the sensors at a specific lighthouse. Detailed insights into the classification task are provided in Section \ref{sec:classification}.

Once we have established this baseline, we proceed to simulate faults within the data. The purpose of this simulation is to examine the impact of these induced irregularities on the performance of the pre-trained ML models. For instance, if a model demonstrates a 90\% accuracy rate with the 'normal' data (baseline), we would expect a noticeable decrease in accuracy when the model is applied to data that exhibits consistent irregular sensor behavior. The methodology behind our simulation of this irregular behavior is further elaborated in Section \ref{sec:drift}.

\subsection{Classification}\label{sec:classification}

The GRAD team records a variety of data from light sensors at lighthouses, including the date and time of each observation, the sun angle, and the operational status of the sensor light (either on or off), which forms the basis of our binary classification problem. To enrich this dataset, we've incorporated several climate-related variables: temperature, dew point, pressure, precipitation, global horizontal irradiance (GHI), diffuse horizontal irradiance (DHI), and Beam Normal Irradiation (BNI). The inclusion of these climate variables is crucial as they could significantly influence the sensor's functionality. For example, intense precipitation could result in darker atmospheric conditions, potentially triggering the lighthouse light.

Using these features, our objective is to train machine learning algorithms to develop models capable of discerning the climate conditions under which a lighthouse light is either on or off. We employ the \verb|sklearn| Python library \cite{pedregosa2011scikit} for implementing algorithms like decision trees, random forest, extreme gradient boosting, and multi-layer perceptron. We standardize all features using \verb|sklearn|'s \verb|StandardScaler|, which normalizes the data by eliminating the mean and scaling to unit variance. The performance of each classifier is evaluated based on accuracy and F1 score, the latter being the harmonic mean of precision and recall.

Once the machine learning algorithms are applied to the historical data, we establish a baseline performance (accuracy and F1 score) for each lighthouse. Subsequently, we analyze how this performance fluctuates when we begin simulating faults. The methodology for simulating these faults is detailed in the following section.

\subsection{Drifting sensor on/off times}\label{sec:drift}

A sensor failure is defined as a malfunction where the sensor does not activate or deactivate the lighthouse light precisely at the critical moments of sunset and sunrise. Such failures can manifest either gradually, with increasing delays in response due to reduced light sensitivity, or abruptly, where the sensor ceases to function entirely. The latter scenario is more straightforward; a completely non-functional sensor means the light remains perpetually on or off, eliminating the need for machine learning analysis.

To focus on gradual sensor failures, we have implemented a `drift' in the timing of the light's activation in the evening and deactivation in the morning. This is based on the premise that a malfunctioning sensor would likely have a delayed response to changing light conditions, resulting in a slower reaction both in the morning (leading to delayed light deactivation) and in the evening (causing delayed light activation).

Gradual photoresistor sensor faults manifest in two distinct ways: first, a decrease in the sensor threshold, where even minimal sunlight can trigger a change in the light source status; and second, an increase in the threshold, necessitating more sunlight to prompt a status change. The decreased threshold fault is particularly critical, as it can lead to the AtoN failing to operate its light source in poor visibility conditions, such as limited sunlight or fog, thereby increasing the risk of accidents. Consequently, our research predominantly addresses faults involving a gradually decreasing sensor threshold.

To accurately simulate these decreasing threshold faults, we employ a drifting operation in our data processing. This involves adding a fixed amount of time to the dataset entry timestamp for events when the light source turns on, while subtracting the same duration for turn-off events. This approach not only modifies the event timestamps but also necessitates the adjustment of related features (such as sun angle and climate data) to align with the updated, 'drifted' times. As a result, each newly created dataset offers a realistic representation of data from a faulty light sensor.

For our experiments, we generate several datasets with varying degrees of drift—specifically, 1, 5, 10, 15, 20, 25, and 30 minutes. The rationale behind creating multiple datasets with different drift intervals is to examine how the algorithms' performance, particularly in terms of classification accuracy and F1 scores, is affected as the degree of time drift increases.

To analyze these potential performance declines, we refer back to the best performing classifier identified in the classification step (see Section \ref{sec:classification}). This classifier serves as the baseline for our drift experiments. By applying the same trained model across different drifted datasets, we anticipate a consistent decline in accuracy and F1 scores as the drift magnitude escalates. Our experiments aim to determine whether all lighthouses' trained ML models exhibit this continuous degradation in accuracy and F1 scores under varying drift conditions.

\section{Experimental setup}\label{sec:setup}
In this section, we present the details of our experimental setup. We first present in Section \ref{sec:data} the datasets used in our experiments. Then, in Section \ref{sec:algos}, we present the machine learning algorithms used in our experiments, along with the hyperparameter tuning process.

\subsection{Datasets}\label{sec:data}
% The experiments for Sections \ref{sec:classification} and \ref{sec:drift} are performed on seven lighthouses' datasets. All lighthouses come from the Trinity House (TH) lighthouse authority. These lighthouses are: Bishop Rock, Eddystone, Godrevy, Lizard, Longships, Trevose, and Wolfrock. We use 3.5 years' worth of data, from June 2017 to December 2021. The timestamps, sun angle, and the sensor light status are obtained by Tritiy House. The climate variables (temperature, dew point, pressure, precipitation, GHI, DHI, and BNI) are obtained by the Copernicus EU Project's CDS and ADS online archives \cite{copernicusAdsCams,copernicusCdsEra5}. Number of observations per dataset varies, depending on how many on/off records exist. This variability is not surprising, because many factors can affect the lighthouse light turning on/off, e.g. cloudy conditions can trigger the light going on, and then when the sky clears, the light will go off, and this can take place on multiple occasions during a day. On average each dataset has around 3,500-4,500 observations.

The experimental work detailed in Sections \ref{sec:classification} and \ref{sec:drift} was conducted using datasets from seven lighthouses, all under the jurisdiction of Trinity House (TH). These lighthouses include Bishop Rock, Eddystone, Godrevy, Lizard, Longships, Trevose, and Wolfrock. Our analysis encompasses 3.5 years of data, spanning from June 2017 to December 2021.

The data components, specifically the timestamps, sun angle, and sensor light status, were sourced from Trinity House. In addition, we incorporated a range of climate variables into our study, including temperature, dew point, pressure, precipitation, Global Horizontal Irradiance (GHI), Diffuse Horizontal Irradiance (DHI), and Beam Normal Irradiance (BNI). These climate variables were obtained from the Copernicus EU Project's Climate Data Store (CDS) and Atmospheric Data Store (ADS) online archives \cite{copernicusAdsCams,copernicusCdsEra5}.

The number of observations in each dataset varies, contingent on the frequency of the light’s on/off cycles. This variation is expected, given that multiple factors can influence the lighthouse light's operation. For instance, cloudy conditions might activate the light, which could then deactivate when the sky clears – a cycle that can occur several times in a single day. On average, each dataset contains approximately 3,500 to 4,500 observations.

\subsection{Machine learning algorithms}\label{sec:algos}
We use python's \verb|sklearn| library to run four machine learning algorithms, namely: decision trees, random forest, extreme gradient boosting (XGBoost), and multi-layer perceptron (MLP). 

To tune each algorithm's hyperparameters, we use 10-fold cross validation, by using \verb|sklearn|'s built-in \verb|GridSearchCV| function. Given that each algorithm is separately tuned for each dataset, the resulted tuned hyperparameters are tuned for each individual dataset. The range of hyperparameter values used for tuning each algorithm are presented in Table \ref{tab:hyperparameters}.

\begin{table*}[htbp]
	\setlength{\tabcolsep}{5.5pt} %% <---------adjust the value here
	%\footnotesize
	\centering
	\caption{ML algorithms hyperparameters range for sklearn. Hyperparameters not mentioned in this table use the default value provided by sklearn.}
	\begin{tabular}{p{3cm}ll}
		\hline
		Algorithm & Hyperparameter &  Value range    \\
		
		\hline
\multirow{1}{*}{Decision tree}
& Criterion & gini, entropy \\
& Max depth & None, 2, 4, 8, 10 \\
& Max no of features & None, sqrt, log2, 0.2, 0.4, 0.6, 0.8 \\
& Splitter & best, random \\
\hline
\multirow{1}{*}{Random forest}
& Max depth & 2, 4, 8, 10\\
& No of estimators & 100, 200, 500\\
& Learning rate & 0.1, 0.01, 0.05\\
\hline
\multirow{1}{*}{XGBoost}
& No of estimators & 100, 200, 500 \\
& Max depth & 6, 9, 12 \\
& Learning rate & 0.1, 0.01, 0.05 \\
\hline
\multirow{1}{*}{MLP}
& Hidden layer sizes & (9), (18), (9,9), (18,9)\\
& Activation & tanh, relu, logistic\\
& Solver & sgd, adam\\
& Alpha & 0.0001, 0.05, 1\\
& Learning rate & constant, adaptive\\
\hline
	\end{tabular}
	\label{tab:hyperparameters}
	
\end{table*}

\section{Results}\label{sec:results}
% As previously explained, the aim of our experiments is to allow us to detect sensor faults as early as possible. In order to do this, we first need to establish a baseline classification performance on historical data for each lighthouse. We present the results from this investigation in Section \ref{sec:classRes}, where we present the accuracy and F1 scores for each classifier per dataset. Then, in Section \ref{sec:driftResults}, we present and discuss how the classification performance is affected when a pre-trained ML model is applied to drifted data. This allows us to investigate if continuous drops exist, which would offer us evidence that ML algorithms can be used to identify faulty sensors.

As previously mentioned, the primary goal of our experiments is to enable the early detection of lighthouse photoresistor sensor faults. To achieve this, we first establish a baseline of classification performance using historical data for each lighthouse. The results of this initial phase are detailed in Section \ref{sec:classRes}, where we report the accuracy and F1 scores for each classifier across the different datasets.

Subsequently, in Section \ref{sec:driftResults}, we delve into the impact of applying pre-trained machine learning models to `drifted' data. This step is crucial for determining whether the classification performance shows continuous declines, a pattern that would indicate the potential of ML algorithms to effectively identify faulty sensors. By analyzing these performance trends, we aim to validate the effectiveness of machine learning techniques in the early detection of sensor malfunctions.

\subsection{Classification results}\label{sec:classRes}
% Table \ref{tab:accuracy} presents the test set accuracy results for the four ML algorithms, over the seven datasets. As we can observe, all algorithms are showing high accuracy, which is mainly around 80-85\%. The only exception is Godrevy, which shows accuracy around 96-97\%. It should be noted such accuracy differences are not important in this particular problem, because our aim is not to find the dataset with the best classification results. These results will act as a baseline, or `normal' behaviour for that particular station. Then, when we start simulating sensor faults (see Section \ref{sec:driftResults}), we can study how the classification results are affected, e.g. if the accuracy drops when the sensor leads to the lighthouse's lights turning on/off with delays. Results are very similar for the F1 score for both the `On' (Table \ref{tab:f1_1}) and the `Off' (Table \ref{tab:f1_0}) classes, where we can again observe values of 80-85\% for all lighthouses apart from Godrevy, which is again around 96-97\%.

Table \ref{tab:accuracy} in our study presents the test set accuracy results for the four machine learning (ML) algorithms across the seven datasets. The results indicate high accuracy levels for all algorithms, generally ranging between 80-85\%. An outlier in this trend is the Godrevy dataset, which exhibits notably higher accuracy, around 96-97\%. However, it's important to emphasize that the primary focus of our investigation is not to pinpoint the dataset with the most accurate classification results. Instead, these figures serve as a baseline or a representation of `normal' behavior for each specific lighthouse.

The significance of these baseline results will become evident when we begin simulating sensor faults, as discussed in Section \ref{sec:driftResults}. This simulation allows us to analyze the impact of faults on classification performance, such as observing how accuracy might decrease when sensor faults cause delays in the lighthouse lights turning on or off. Results are very similar for the F1 score for both the `On' (Table \ref{tab:f1_1}) and the `Off' (Table \ref{tab:f1_0}) classes, where we can again observe values of 80-85\% for all lighthouses apart from Godrevy, which is again around 96-97\%.

\begin{table*}[ht!]
	
	\setlength{\tabcolsep}{3.5pt} %% <---------adjust the value here
	%\footnotesize
	\centering
	\caption{\% Accuracy for each ML algorithm over seven lighthouses.}
	\begin{tabular}{lcccccccc}
		\hline
	Dataset & DT	& RF	& XGBoost	& MLP\\
 \hline
Bishop Rock	& 83.30\% &	83.76\% &	83.85\% &	84.59\%\\
Eddystone	& 85.80\% &	85.02\% &	83.46\% &	85.41\%\\
Godrevy	& 96.27\% &	97.06\% &	97.06\% &	97.18\%\\
Lizard	& 82.98\% &	83.32\% &	82.64\% &	84.00\%\\
Longships	& 82.99\% &	84.06\% &	82.46\% &	83.70\%\\
Trevose	& 81.27\% &	82.94\% &	82.59\% &	84.43\%\\
Wolfrock	& 85.18\% &	86.75\% &	85.79\% &	85.35\%\\
\hline
	\end{tabular}
	\label{tab:accuracy}
	
\end{table*}

\begin{table*}[ht!]
	
	\setlength{\tabcolsep}{3.5pt} %% <---------adjust the value here
	%\footnotesize
	\centering
	\caption{\% F1 score for class `On' for each ML algorithm over seven lighthouses.}
	\begin{tabular}{lcccccccc}
		\hline
	Dataset & DT	& RF	& XGBoost	& MLP\\
 \hline
Bishop Rock &	84.88\% &	83.32\% &	84.85\% &	84.86\%\\
Eddystone &	83.10\% &	82.18\% &	82.97\% &	84.97\%\\
Godrevy &	96.19\% &	97.04\% &	97.03\% &	97.17\%\\
Lizard &	83.55\% &	82.69\% &	83.00\% &	83.85\%\\
Longships &	84.54\% &	84.43\% &	83.72\% &	84.62\%\\
Trevose &	81.59\% &	81.42\% &	82.69\% &	84.15\%\\
Wolfrock &	85.74\% &	85.55\% &	85.35\% &	84.97\%\\
\hline
	\end{tabular}
	\label{tab:f1_1}
	
\end{table*}

\begin{table*}[ht!]
	
	\setlength{\tabcolsep}{3.5pt} %% <---------adjust the value here
	%\footnotesize
	\centering
	\caption{\% F1 score for class `Off' for each ML algorithm over seven lighthouses.}
	\begin{tabular}{lcccccccc}
		\hline
	Dataset & DT	& RF	& XGBoost	& MLP\\
 \hline
Bishop Rock &	81.35\% &	84.18\% &	82.71\% &	84.30\%\\
Eddystone &	87.75\% &	87.08\% &	83.93\% &	85.82\%\\
Godrevy &	96.35\% &	97.09\% &	97.09\% &	97.18\%\\
Lizard &	82.36\% &	83.91\% &	82.26\% &	84.15\%\\
Longships &	81.10\% &	83.67\% &	81.00\% &	82.67\%\\
Trevose &	80.93\% &	84.23\% &	82.48\% &	84.70\%\\
Wolfrock &	84.57\% &	87.77\% &	86.20\% &	85.71\%\\
\hline
	\end{tabular}
	\label{tab:f1_0}
	
\end{table*}

% To identify the ML algorithm that performed best, we run the non-parametric Friedman test, under the null hypothesis that all algorithms' observations come from the same continuous distribution. Table \ref{tab:friedman} presents the average rank of each algorithm for accuracy (left), F1 for `On' (middle), and F1 for `Off' (right). In addition to the average ranks, the table also presents the adjusted p-value according to the two stage Benjamin/Hochberg post hoc test. Statistical significance occurs when the p-value is below 0.05 (5\% significance level). As we can observe, MLP ranks first across all three metrics. Although it does not statistically outperform the other ML algorithms, it shows significance close to 10\% level when compared to XGBoost and DT for accuracy and F1 (`Off') (p-values 0.13 and 0.11, respectively). Even though MLP does not show statistical significance against the other algorithms, this is not of concern, because as we mentioned earlier the goal of this experiment was to to act as a baseline for the upcoming drifting experiments. The most important observation is that MLP ranks first across both accuracy and the two F1 scores. As a result, we will use the MLP for the next set of experiments.

To determine the most effective machine learning (ML) algorithm among those tested, we employed the non-parametric Friedman test. This test operates under the null hypothesis that all algorithms' observations are drawn from the same continuous distribution. The results of this test are displayed in Table \ref{tab:friedman}, which shows the average rank of each algorithm in terms of accuracy (left column), F1 score for the 'On' class (middle column), and F1 score for the 'Off' class (right column). Additionally, the table includes the adjusted p-values according to the two-stage Benjamin/Hochberg post hoc test, a method for controlling the false discovery rate.

Statistical significance in this context is defined by a p-value below 0.05, indicative of a 5\% significance level. Our observations reveal that the Multi-Layer Perceptron (MLP) algorithm consistently ranks first across all three metrics. While it does not achieve statistical significance over the other ML algorithms, it does approach significance at the 10\% level when compared to Extreme Gradient Boosting (XGBoost) and Decision Trees (DT) in terms of accuracy and F1 score for the `Off' class, with p-values of 0.13 and 0.11, respectively.

Although the MLP doesn't exhibit statistical significance in comparison to the other algorithms, this is not a major concern for our study. As previously mentioned, the primary objective of this phase of the experiment was to establish a baseline for the subsequent drift experiments, rather than to conclusively identify the superior algorithm. The key takeaway is that the MLP ranks first in both accuracy and the two F1 scores. Therefore, we will utilize the MLP for the forthcoming set of experiments.

\begin{table*}[ht!]
	
	\setlength{\tabcolsep}{2.5pt} %% <---------adjust the value here
	%\footnotesize
	\centering
	\caption{From left to right: Friedman tests for Accuracy, F1 for `On' class, and F1 for `Off' class.}
	\begin{tabular}{lcc|lcc|lcc}
		\hline
Algorithm &	Rank &	Adj. p-value & Algorithm &	Rank &	Adj. p-value & Algorithm &	Rank &	Adj. p-value\\
\hline
MLP (c) &	1.57 &	- &	MLP (c) &	1.57 &	-	 &MLP (c) &	1.71 &	-\\
RF &	2.00 &	0.58 &	DT &	2.14 &	0.51 &	RF &	1.85 &	0.84\\
XGBoost &	3.00 &	0.13 &	XGBoost &	3.00 &	0.17 &	XGBoost &	3.14 &	0.11\\
DT &	3.14 &	0.13 &	RF &	3.28 &	0.15 &	DT &	3.28 &	0.11\\
\hline
	\end{tabular}
	\label{tab:friedman}
	
\end{table*}

\subsection{Drifting results}\label{sec:driftResults}
% As explained in Section \ref{sec:drift}, we simulate errors in the sensors, which will lead to delays of lighthouse lights turning on and off. These delays are 1, 5, 10, 15, 20, 25, and 30 minutes. Thus, seven new datasets are created for each lighthouse, where each dataset has drifted the logged observations by one of the aforementioned times. The purpose of these experiments is to investigate if the classification performance of the best performing MLP model, which was trained in Section \ref{sec:classRes}, will drop as drifting increases. Our hypothesis is that since a model has been trained on a specific set of data, its accuracy and F1 score will drop continuously, as drifting times increase.

As detailed in Section \ref{sec:drift}, we simulate sensor errors that introduce delays in the activation and deactivation of lighthouse lights. These delays are set at 1, 5, 10, 15, 20, 25, and 30 minutes. Consequently, we generate seven distinct datasets for each lighthouse, each one applying a drift of the recorded observations by one of the specified time intervals. The primary objective of these experiments is to assess whether the classification performance of the top-performing MLP model, trained as discussed in Section \ref{sec:classRes}, exhibits a decline as the duration of drifting increases. Our working hypothesis posits that, given the model's training on a specific dataset, its accuracy and F1 score will progressively decrease with longer drift duration.

\begin{figure}[h!]
  \centering
    \includegraphics[width=\linewidth]{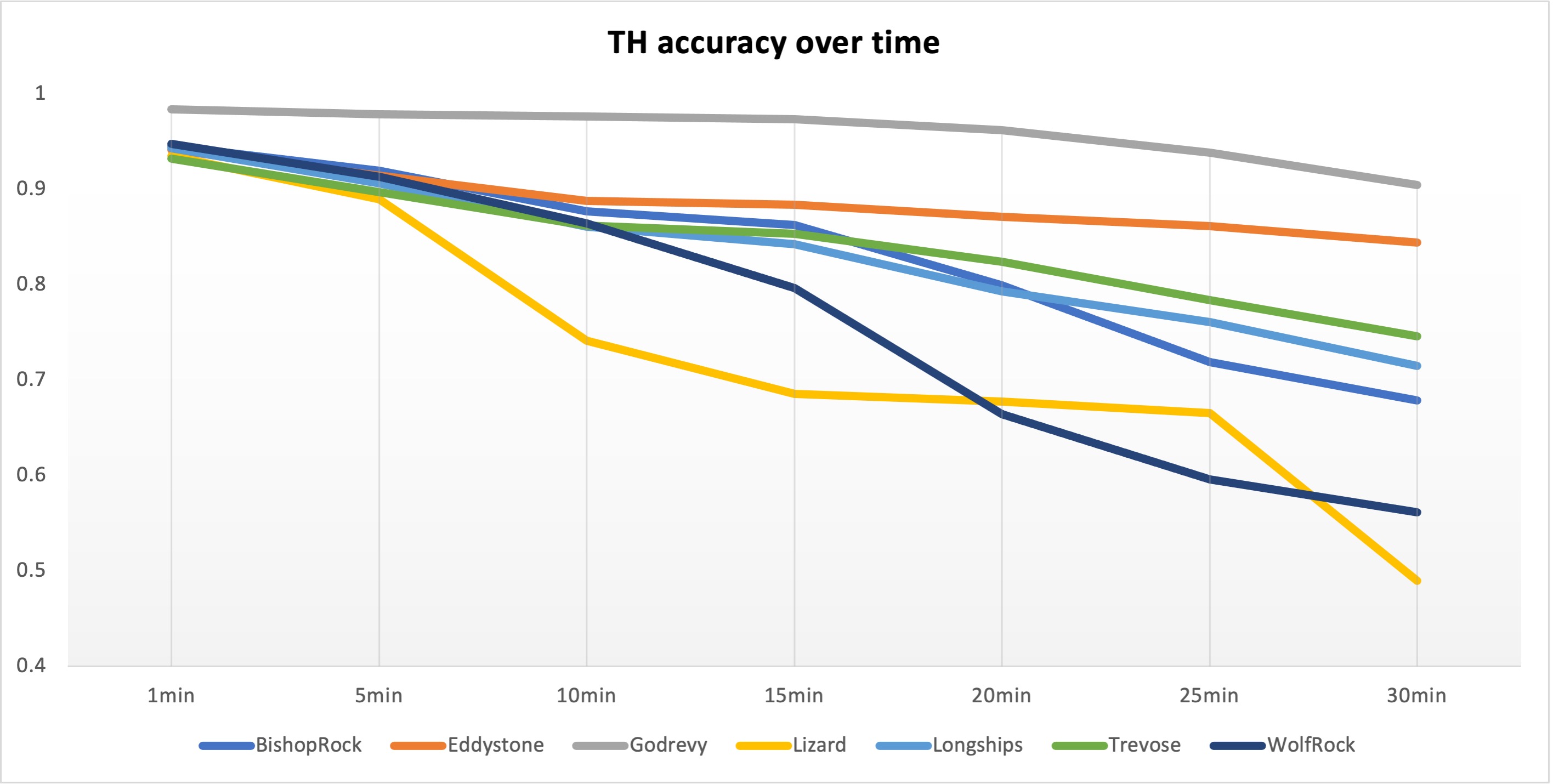}
  \caption{Accuracy performance over different drifted times.}
  \label{fig:drifted_accuracy}
\end{figure}

\begin{figure}[h!]
  \centering
    \includegraphics[width=\linewidth]{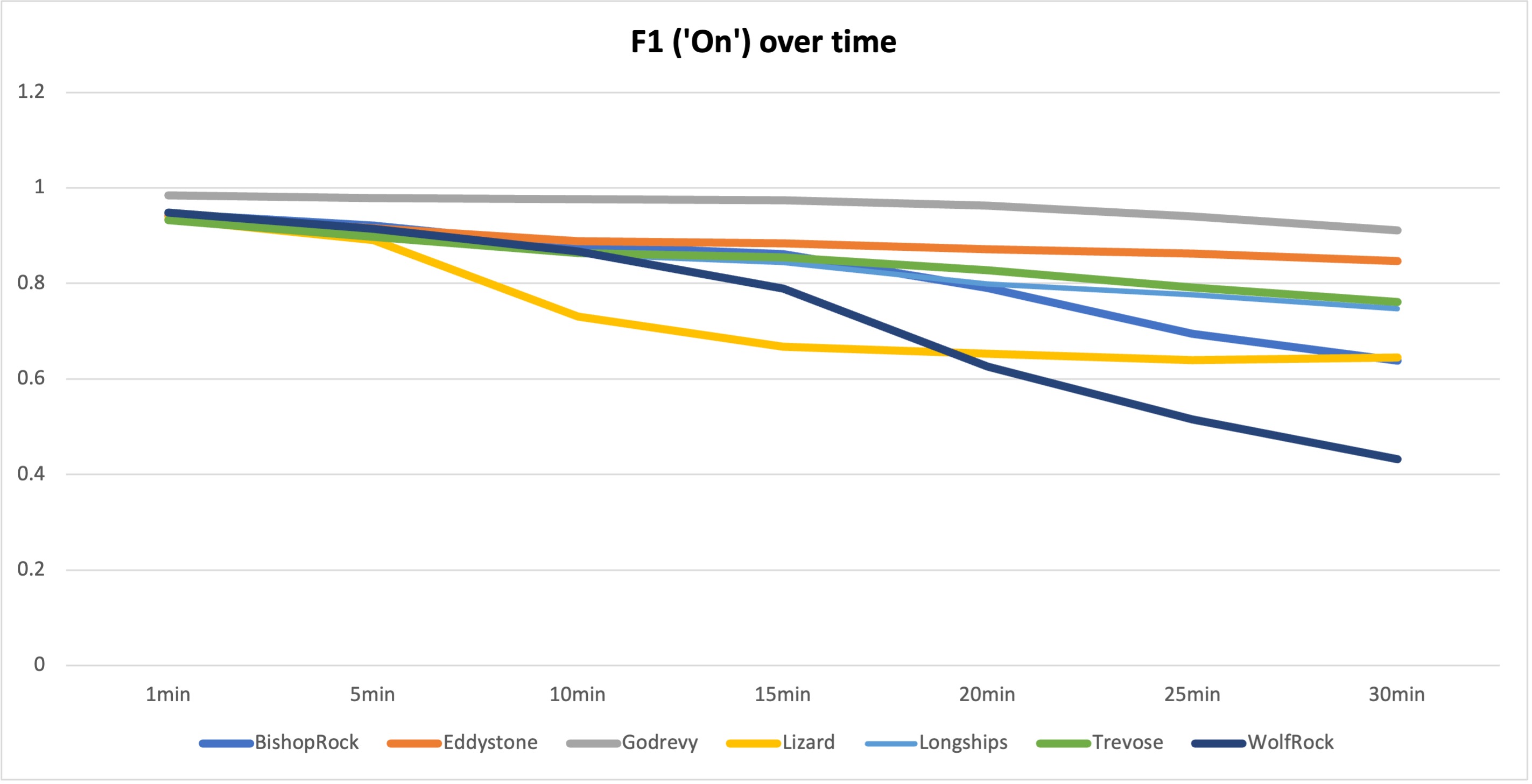}
  \caption{F1 (`On') performance over different drifted times.}
  \label{fig:drifted_f11}
\end{figure}

\begin{figure}[h!]
  \centering
    \includegraphics[width=\linewidth]{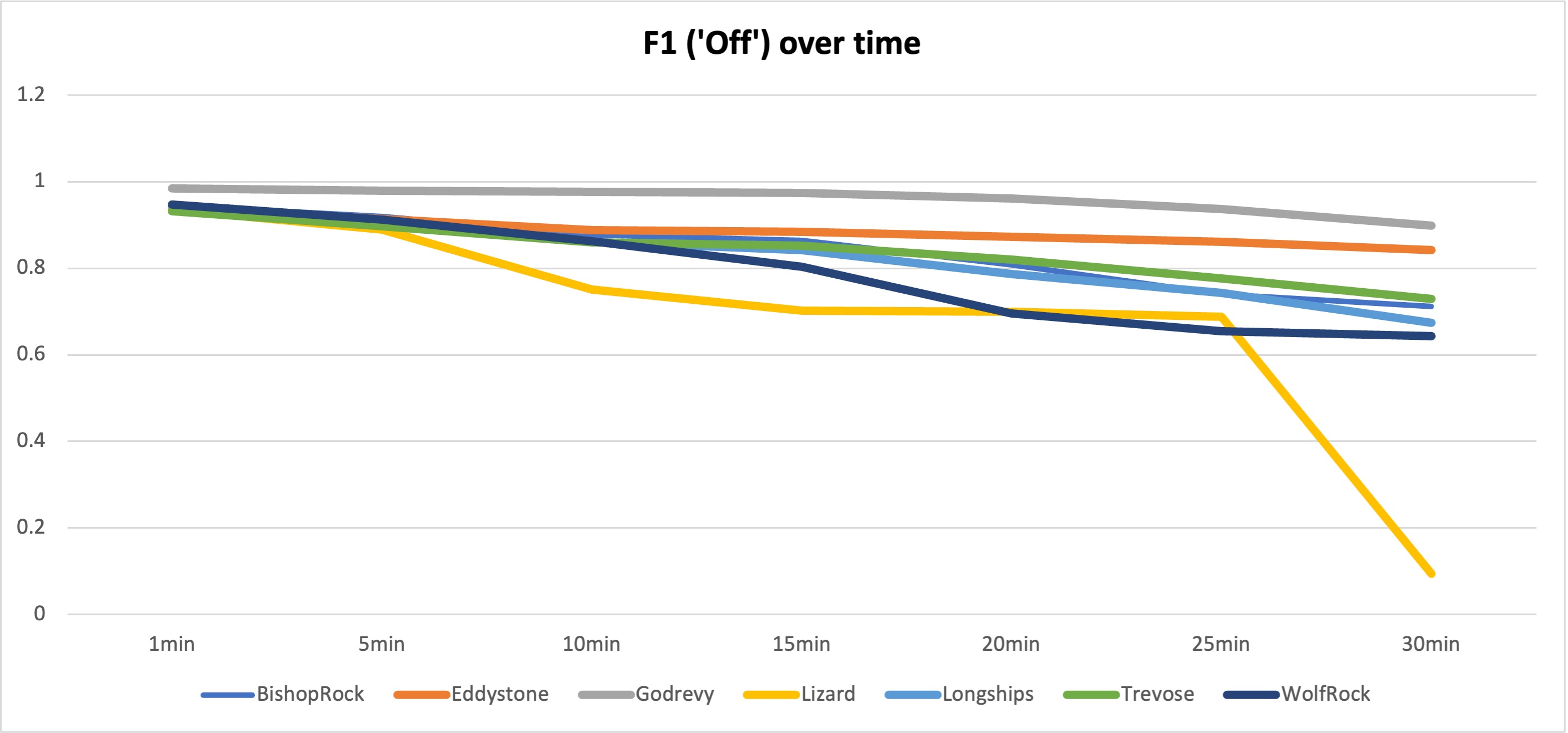}
  \caption{F1 (`Off') performance over different drifted times.}
  \label{fig:drifted_f10}
\end{figure}

% Our hypothesis seems to be confirmed, as we can observe from Figs. \ref{fig:drifted_accuracy}-\ref{fig:drifted_f10}, where we present the accuracy and F1 scores. As we can see, there is a continuous decrease in all metrics across all datasets. We can also observe large drops in accuracy from the 5 or the 10-minute mark (with the exception of Godrevy, which experiences a much slower drop in accuracy). On average, the smaller accuracy decrease is by Godrevy at 7.88\%, and the larger is by Lizard at 44.52\%. We can observe similar continuous decreases in the F1 scores in Figures \ref{fig:drifted_f11} and \ref{fig:drifted_f10}. Table \ref{tab:avgDrops} summarises the mean decreases in accuracy and F1 over the different drifting periods for each dataset, which as we can see ranges from 7.88\% (Godrevy) to 44.52\% (Lizard) for accuracy; from -7.27\% (Godrevy) to -51.64\% (Wolfrock) for F1 (`On'); and from -8.58\% (Godrevy) to -84.09\% (Lizard) for F1 (`Off').

Our initial hypothesis appears to be confirmed, as evidenced by the trends depicted in Figures \ref{fig:drifted_accuracy} to \ref{fig:drifted_f10}, which illustrate the accuracy and F1 scores. These visualizations reveal a consistent decline in all performance metrics across the various datasets. Notably, significant decreases in accuracy become evident at the 5 or 10-minute mark, with the exception of Godrevy, which exhibits a more gradual decline in accuracy. On average, the smallest accuracy decrease is observed in Godrevy at 7.88\%, while the most substantial decrease occurs in Lizard at 44.52\%. Similarly, Figures \ref{fig:drifted_f11} and \ref{fig:drifted_f10} illustrate analogous continuous decreases in the F1 scores.

Table \ref{tab:avgDrops} provides a summary of the average declines in accuracy and F1 scores across different drifting periods for each dataset. These averages range from 7.88\% (Godrevy) to 44.52\% (Lizard) for accuracy, from -7.27\% (Godrevy) to -51.64\% (Wolfrock) for F1 (On'), and from -8.58\% (Godrevy) to -84.09\% (Lizard) for F1 (Off').

\begin{table*}[ht!]
	
	\setlength{\tabcolsep}{5.5pt} %% <---------adjust the value here
	%\footnotesize
	\centering
	\caption{Average accuracy decrease over the different drifting times.}
	\begin{tabular}{lccc}
		\hline
Dataset &	Mean Acc Decrease & Mean F1 (`On') Decrease & Mean F1 (`Off') Decrease\\
\hline
Bishop Rock &	-26.65\% & -30.72\% & -23.39\%\\
Eddystone &	-9.66\% & -9.44\% & -9.90\%\\
Godrevy &	-7.88\% & -7.27\% & -8.58\%\\
Lizard &	-44.52\% & -29.01\% & -84.09\%\\
Longships &	-22.81\% & -19.57\% & -27.00\%\\
Trevose &	-18.65\% & -17.13\% & -20.37\%\\
Wolfrock &	-38.60\% & -51.64\% & -30.41\%\\
\hline
	\end{tabular}
	\label{tab:avgDrops}
	
\end{table*}

% Another interesting observation we can make is on the timing of the accuracy drops. As we can see, the accuracy drop can happen as quickly as within 5 minutes. By the time we've reached the 15-20 minute mark, all lighthouses' accuracy and F1 scores have fallen by at least 5-10\%. Another important observation is the continuous decrease that is observed as the time drift increases. This is very important, because even if in some cases the accuracy and F1 scores decreases might happen at a slower rate (e.g. Godrevy), there is a continuous downward trend across all three metrics. Thus, the fault can be detected overtime.

% The above results seem to confirm the hypothesis that when applying a pre-trained ML model to drifted data, the accuracy and F1 scores will decrease overtime. This thus establishes that machine learning algorithms can be used to detect faults on datasets that haven't shown any faults in the past, and therefore models cannot be directly trained to learn when such faults occur. 

Additional noteworthy observations can be made regarding the timing of accuracy drops. It is evident that accuracy can deteriorate rapidly, with declines occurring within as little as 5 minutes. By the time the 15-20 minute mark is reached, all lighthouses' accuracy and F1 scores have experienced decreases of at least 5-10\%. Furthermore, there is a crucial observation of a continuous decline as the time drift increases. This consistency is significant because, even in cases where accuracy and F1 scores decrease more gradually (e.g., Godrevy), there is an overarching downward trend across all three metrics. Consequently, faults can be detected progressively over time.

The results presented here confirm the hypothesis that applying a pre-trained ML model to drifted data leads to a decrease in accuracy and F1 scores over time. This finding establishes that machine learning algorithms can effectively detect faults in datasets that have not previously exhibited faults, and, as a result, models can be employed to identify when such faults occur.

\section{Conclusion}\label{sec:conclusion}
% To conclude, this article discussed the problem of fault detection for lighthouse light sensors. We explained that as there are no historical data to allow a machine learning algorithm to learn and predict faults, an alternative could be to simulate faults and detect them as early as possible. Our experiments showed that machine learning can detect drifts in sensor behaviours as early as 10-15 minutes. We also found that over time, there is a continuous decrease in machine learning accuracy and F1 score, which can be an effective way of early warning for detecting sensor faults. 

In conclusion, this article addresses the challenge of fault detection in lighthouse light sensors. Given the absence of historical data to train machine learning models for fault prediction, the approach of simulating faults and detecting them early is explored. The experiments demonstrate that machine learning can detect sensor behavior drifts within 10-15 minutes. Moreover, it is observed that accuracy and F1 scores continuously decrease over time, serving as an effective early warning mechanism for sensor fault detection. 

Future work can focus on trying to train ML algorithms on combined lighthouse stations' data. At the current paper, ML algorithms were trained per station. However, this can lead to difficulties in maintaining multiple models. An alternative could be to combine all lighthouse data and apply ML algorithms to create a single, `global', trained model

% Future work can take the following directions: (i) reduce the early warning to less than 10-15 minutes, (ii) fault detection on new lighthouse deployments, where no historical data is available at all, not only for faults, but in general, and (iii) investigate advantages from using more climate features that might be affecting a sensor's behaviour.

\section*{Acknowledgements}
This work was funded by the GLA Research and Development (GRAD) division, of the General Lighthouse Authorities of the UK and Ireland.

\bibliographystyle{IEEEtran}
\bibliography{IEEEabrv,sample}
\end{document}